# An Exploratory Study of Hierarchical Fuzzy Systems Approach in A Recommendation System

**Tajul Rosli Razak[1*], Iman Hazwam Abd Halim[2], Muhammad Nabil Fikri Jamaludin[3], Mohammad Hafiz Ismail[4], Shukor Sanim Mohd Fauzi[5]**

[1,2,3,4,5] *Faculty of Computer and Mathematical Science, Universiti Teknologi MARA, 02600 Arau, Perlis, Malaysia*

*Authors' Email Address:* *[*1]tajulrosli@uitm.edu.my, [2]hazwam688@uitm.edu.my, [3]nabilfikri@uitm.edu.my, [4]mohammadhafiz@uitm.edu.my, [5]shukorsanim@uitm.edu.my,*

## ABSTRACT

*Recommendation system or also known as a recommender system is a tool to help the user in providing a suggestion of a specific dilemma. Thus, recently, the interest in developing a recommendation system in many fields has increased. Fuzzy Logic system (FLSs) is one of the approaches that can be used to model the recommendation systems as it can deal with uncertainty and imprecise information. However, one of the fundamental issues in FLS is the problem of the curse of dimensionality. That is, the number of rules in FLSs is increasing exponentially with the number of input variables. One effective way to overcome this problem is by using Hierarchical Fuzzy System (HFSs). This paper aims to explore the use of HFSs for Recommendation system. Specifically, we are interested in exploring and comparing the HFS and FLS for the Career path recommendation system (CPRS) based on four key criteria, namely topology, the number of rules, the rules structures and interpretability. The findings suggested that the HFS has advantages over FLS towards improving the interpretability models, in the context of a recommendation system example. This study contributes to providing an insight into the development of interpretable HFSs in the Recommendation systems.*

## INTRODUCTION

Recommendation systems are software tools and techniques, providing suggestions for items to be of use to a user (Ricci et al., 2011). The suggestions may be related to various decision-making processes, such as what items to buy, what music to listen to, or what online news to read. The recommendation systems are commonly used over the Internet to guide customers to find the products or services that best fit with their personal preferences. Also, as discussed in (Burke, 2002), recommendation systems may also represent user preferences to suggest items to purchase or examine.

The field of FLSs has been making rapid progress in recent years. There has been an increasing number of works in many areas such as science, manufacturing, business and also in the recommendation systems (Razak et al., 2014) for decision making. Many researchers (González et al., 2015; Samuel et al., 2013) have used fuzzy systems as a tool for controlling and modelling in many fields, proving it to be a useful technology. This is mainly due to the flexibility and ease of which knowledge can be expressed
through fuzzy rules as well as some theoretical results in these fields (Zadeh, 1965).

One of the strengths of FLSs is their interpretability, particularly in applications such as knowledge extraction and decision support (Mikut et al., 2005; D. Nauck & Kruse, 1999). However, key challenges remain around FLS interpretability, including the *curse of dimensionality*: the number of required rules commonly increases exponentially with the number of input variables (Zhou et al., 2009). This challenge is also known as rule explosion, which may reduce the transparency and interpretability of FLSs (S. Jin & Peng, 2015). One effective way to deal with this problem is through the use of a special type of FLS, namely HFSs (Raju et al., 1991).

HFSs were introduced as an approach to overcome the *curse of dimensionality*, which arises in conventional FLSs (Raju et al., 1991). In HFSs, the original FLSs are decomposed into a series of low-dimensional FLSs—fuzzy logic subsystems. Moreover, the rules in HFSs commonly have antecedents with fewer variables than the rules in FLSs with equivalent function, since the number of input variables of each subsystem is lower (Benítez & Casillas, 2013; Salgado, 2008). Thus, HFSs tend to reduce rule explosion, thus minimizing complexity, and improving model interpretability.

Based on these advantages, in this paper, we intend to explore and examine the use of HFSs in practice, particularly in the recommendation systems. Specifically, we will compare and observe the HFS with the flat FLSs based on four key criteria, namely *topology, the number of rules, the rules structures* and *interpretability*. An initial study of modelling the recommendation systems using the flat FLS was previously proposed by the authors (Razak et al., 2014).

This paper is organised as follows; The first section discusses the background to FLS, interpretability, recommendation system and career selection. This is followed by the second section that introduces the hierarchical fuzzy systems approach. The third section demonstrates the HFS approach with the real-world example, i.e., career path recommendation system in order to explore and compare the features with the flat FLS. Finally, the last two sections present the discussion, conclusion and future works.

## BACKGROUND

In this section, we briefly provide background in respect to fuzzy logic systems, interpretability in FLSs, recommendation systems and career selections.

### Fuzzy Logic Systems

Fuzzy logic systems (FLSs) are one of the currently used techniques for modelling non-linear, uncertain and complex systems. An essential characteristic of FLSs is the partitioning of the space of system variables into fuzzy regions using fuzzy sets (Zadeh, 1965). In each region, the characteristics of the system can be described merely using a rule. Generally, an FLS consists of a rule base with rules associated with particular regions, where the information available is transparent and easily readable. This characteristic of fuzzy systems has been employed in many fields including medical (Mendez et al., 2018; Razak et al., 2013), engineering (Sahoo et al., 2018), agriculture (Razak et al., 2012), decision support (Samuel et al., 2013), pattern recognition (González et al., 2015), recommendation system (Ismail et al., 2015) and others.

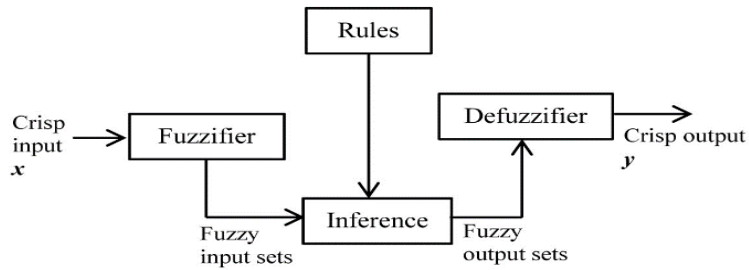

**Figure 1**: **Fuzzy Logic Systems. Adapted from** (Mendel, 2000).

In general, the main steps, as shown in Figure 1 performed in the FLS include: (i) the fuzzification component transforms each crisp input variable into a membership grade based on the membership functions defined; (ii) the inference engine then conducts the fuzzy reasoning process by applying the appropriate fuzzy operators in order to obtain the fuzzy set to be accumulated in the output variable; and (iii) the defuzzifier transforms the fuzzy output into a crisp output by applying a specific defuzzification method.

As mentioned earlier, one of the strengths of FLSs is their interpretability. In contrast, neural networks are viewed as black boxes because mathematical formulae set the mapping between inputs and outputs. Black-box modelling can simulate a real-world system reliably and precisely, but the model structure and parameters usually give no explicit explanation about the system behaviour (Zhou & Gan, 2008). Conversely, FLSs can be seen as grey boxes in the sense that every element of the whole system can be checked for plausibility (Alonso et al., 2009), in which relationships between input and output variables are established in terms of a fuzzy logic-based descriptive language. Also, FLSs consist of linguistic variables (Zadeh, 1975) and linguistic rules (Mamdani, 1977) which are easy to interpret by the user (Setnes et al., 1998) because they are quite close to expert natural language.

**Interpretability of FLSs**

Interpretability indicates how easily an FLS can be understood by human beings (Y. Jin, 2000). In recent years, the interest of researchers in obtaining more interpretable fuzzy models has increased. However, the choice of an appropriate interpretability measure is still an open discussion due to its subjective nature and a large number of factors involved. Substantial research on interpretability measures (Alonso et al., 2006; Gacto et al., 2011; D. D. Nauck, 2003; Zhou & Gan, 2008) proposed interpretability indices for FLSs. The most common interpretability indices are Nauck index and Fuzzy index. An FLS model is said to be less interpretable when its Nauck and Fuzzy index is closer to 0 and more interpretable when Nauck and Fuzzy index is closer to 1. Detail explanation of these Nauck and Fuzzy index can be seen in (Razak et al., 2017).

A further study by (Razak et al., 2017) proposed an initial index, HFSi from the extension of two most common FLS interpretability indices, namely Nauck and Fuzzy index. The HFSi index intends to measure interpretability of HFSs, with a specific focus on the complex structure of HFSs such as having multiple layers, subsystems and different topologies. The HFSi is computed as follows:

$$HFSi = \sum_{i=1}^{n} \left( l_i \sum_{j=1}^{m_i} E_{ij}/m_i \right) \quad (1)$$

where $E_{ij}$ is for example the Nauck (N) or Fuzzy (F) index of a subsystem $j$ at layer $i$, $l_i$ is the associated weight to the layer $i$ of the HFSs, $m_i$ is the number of subsystems located at the layer $i$, and $n$ is the number of layers. Note that (1) returns the original FLS index when applied to a flat FLS. Likewise, an HFS model is less interpretable when the HFSi is close to 0 and more interpretable when HFSi is close to 1.

**Recommendation Systems**

According to (Ricci et al., 2011), a recommender system is a software application and approach which help users find better choices from among a massive list of alternatives. This system can help the user by recommending related to various decision-making techniques. The examples of decision-making techniques that widely used are what product to purchase, what book to read, and so on. Users will easily decide by using a recommender system.

The recommender system is usually the one which provides suggestions and recommendations to users when they are facing different choices in making a decision. Burke in (Burke, 2002) defines any system which can provide individualized recommendations or have the ability to help users in a personalized manner to identify interesting information on things in a big space of possible alternatives. Also, as stated by (Jannach & Friedrich, 2013), the recommendation system is useful to assists users to match items if there is ease of details as well as act like sales assistance, guidance, advisory and others.

Other studies show that several applications give recommendations to users (Segaran, 2007). For example, a book recommendation for online shopping, suggest attractive websites or help the user search for music and movies. This recommender system shows the user the way to build a system to find users that share an absence to create a recommendation system based on the things will also be well-liked by other users.

**Career Selection**

Career selection is one in every of several vital decision's students can create in deciding plans. One area concerning students career development is career choices that will relate to one's career decision making (Talib & Aun, 2009).

As discussed by (Moy & Lee, 2002), career characteristic is essential elements that influence the selection of career among university students. The career characteristic which might be relevant to students in choosing their career is usually classified into three groups which include the career itself, compensation or security as well as the organisation or work environment.

Furthermore, as explained by (Talib & Aun, 2009), career guidance offered in university should fulfil the technical information needs of university students at different levels of their career development. It is necessary to deliver career guidance in several ways such as courses, training and seminars that offer group experiences in future career planning and group or individual counselling activities.

## HIERARCHICAL FUZZY SYSTEMS APPROACH

Hierarchical fuzzy systems (HFSs) are defined by composing the input variables into a collection of low-dimensional fuzzy logic subsystems (Raju et al., 1991). Also, HFSs can be illustrated as a cascade structure where the output of each layer is considered as an input to the following layer, as shown in Figure 3. Moreover, a system that goes from one layer, as shown in Figure 2 to two layers, as in Figure 3 has fewer rules than the one in one layer (Razak et al., 2017).

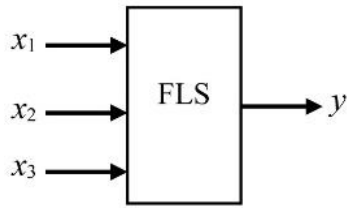 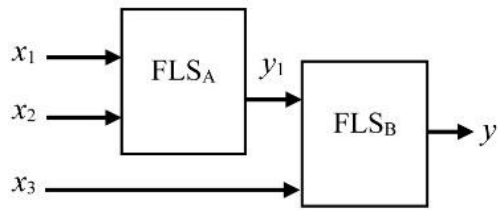

**Figure 2: FLS topology**     **Figure 3: HFS topology**

The most extreme reduction of rules will be if the structure of the HFS has two input variables for each low dimensional FLS and has $(n-1)$ layers — serial structure, where $n$ is the total number of input variables. If we define $m$ fuzzy sets for each input variable, including the intermediate output variables $y_1, \ldots, y_{n-2}$, the total number of rules ($R_{HFS}$) is a linear function (Wang, 1998) of the number of input variables $n$ and can be expressed as:

$$R_{HFS} = (n-1)m^2 \qquad (2)$$

In conventional FLSs, the number of rules increases exponentially with the number of input variables (Lee et al., 2003). Suppose there are $n$ input variables and $m$ fuzzy sets for each input variable, then the number of rules ($R_{FLS}$) needed to construct a complete fuzzy system with fully specified rule base (using the "AND" logical connective) can be expressed as:

$$R_{FLS} = m^2 \qquad (3)$$

From these equations (2) and (3), it is clear that the total number of rules in the FLSs ($R_{FLS}$) is always greater than or equal than the equivalent HFSs ($R_{HFS}$). For example, Figure 2 and Figure 3 show an FLS and HFS with 3 input variables ($n = 3$) and, assuming that 3 fuzzy sets ($m = 3$) are defined for each input variable, the total number of rules for this FLS is $R_{FLS} = m^n = 3 \times 3 = 27$ whereas for the HFS, the total number of rules is $R_{HFS} = (n-1)m^2 = (3-1)3^2 = 18$. It is clear that the total number of rules is always lower or equal when employing an HFS. Consequently, HFS seems to be a practical approach to reduce the model complexity and thus to improve model interpretability.

Therefore, HFSs have been shown to have the potential to improve the interpretability of FLSs (Magdalena, 2019; Razak et al., 2017, 2018). Thus, in this paper, we aim to explore the advantages of HFSs and put forward to apply in the recommendation system and will be demonstrated in the next section. To date, the leveraging of HFSs has not been discussed in details, particularly in the recommendation system.

## EXPERIMENTS AND RESULTS

In this experiment, we use a case study of the career path recommendation system (CPRS) that has been proposed in (Razak et al., 2014), in order to demonstrate the use of HFSs for recommendation system. The CPRS is a recommendation system that provides direction and guidance to students in choosing their career via skills assessment that is based on multiple choice question technique or item rating. The CPRS employed fuzzy logic techniques in order to mapping the skills assessment input to the suggestion career for university students.

**Table 1: A set of questions in skill assessment that used in the CPRS particularly for recommendation of the web programmer career. Adapted from (Razak, Hashim, Noor, Halim, & Shamsul, 2014).**

| Questions | Skills | Scale |
|---|---|---|
| Q1 | Design and develop a web base | 0-10 |
| Q2 | Handle whole web project from start to roll-out | 0-10 |
| Q3 | Skill and knowledge in PHP, HTML, CSS, Javascript and MySQL | 0-10 |
| Q4 | Good in problem solving, communication interpersonal and organization skills | 0-10 |
| Q5 | Up to date with latest web technology trends and programming techniques | 0-10 |

Table 1 shows a set of questions in skill assessment used in CPRS that are mainly to suggest whether the student is fit to choose a Web Programmer for their career path. Note that only a set of skill questions for the career path of a Web Programmer is used in this experiment to demonstrate the use of HFSs. Then, all input, intermediate and output variables are model with three membership functions with linguistic terms is {*Weak, Medium, Good*} and {*No, Maybe, Yes*} respectively, as shown in Table 2. The example of membership functions for input *Q1* can be seen in Figure 4.

**Table 2: Linguistic term for each input, intermediate and output variables for the CPRS.**

| Linguistic variables | Linguistic Term |
|---|---|
| Inputs: | |
| Q1 | {Weak, Medium, Good} |
| Q2 | {Weak, Medium, Good} |
| Q3 | {Weak, Medium, Good} |
| Q4 | {Weak, Medium, Good} |
| Q5 | {Weak, Medium, Good} |
| | |
| Intermediate: | |
| Comb_skill 1 | {Weak, Medium, Good} |
| Comb_skill 2 | {Weak, Medium, Good} |
| Comb_skill 3 | {Weak, Medium, Good} |
| | |
| Output: | |
| Web programmer | {No, Maybe Yes} |

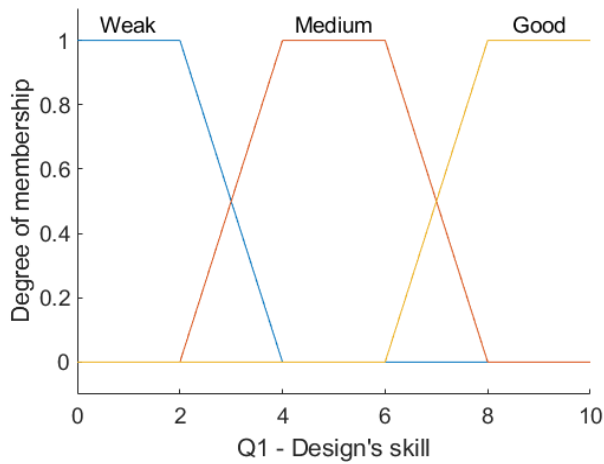

**Figure 4.** Example of membership functions of input Q1 – Design's skill.

This experiment aims to explore and compare the HFSs with the flat FLSs in the CPRS. Usually, most of people tend to compare two systems based on their 'accuracy' in order to choose the best of two. However, beyond the accuracy, in this experiment, the exploration and comparison process is based on the other criteria that include *topology, the number of rules, the rules structure* and *interpretability*.

## Topology

The first criterion that we intend to explore is topology. The topology of FLS and HFS for the CPRS can be seen in Figures 5 and 6, respectively. For the case of FLS, the topology is just a simple to follow as only show a mapping of IO (input-output) connection with single subsystem FLS, as shown in Figure 5. Conversely, for the case of HFS, the topology is produced from the decomposition of the flat FLS (as in Figure 5). Consequently, the topology of the HFS consists of multiple subsystems of FLS and layers, as shown in Figure 6. Also, HFS topology provides additional information such that '*intermediate output*' variables, namely *comb_skill 1, comb_skill 2* and *comb_skill 3,* which are an essential component in mapping the IO connection of HFS.

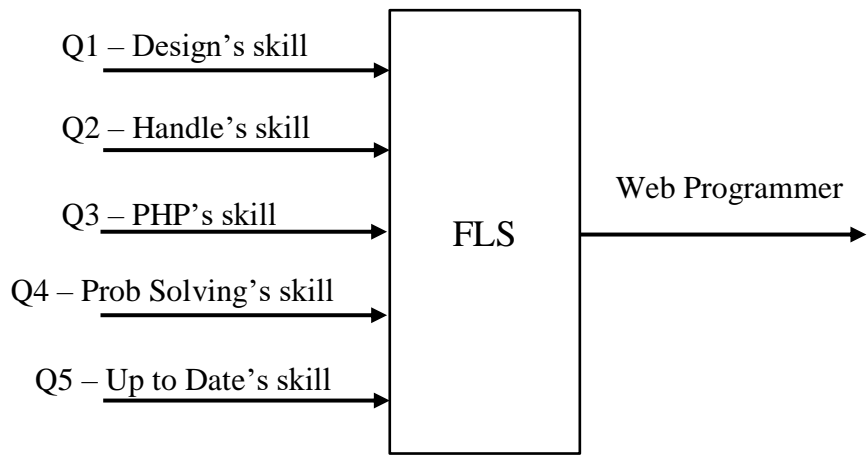

**Figure 5.** Career Path Recommendation System (CPRS) – FLS topology

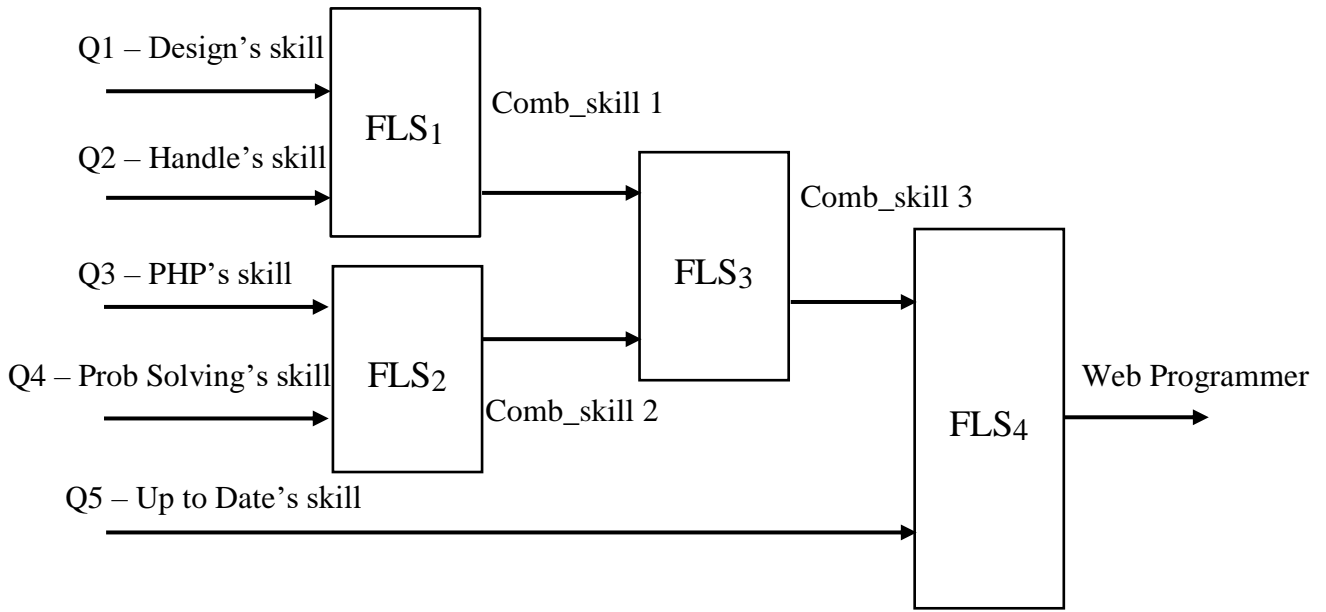

**Figure 6.** Career Path Recommendation System (CPRS) – HFS topology

Despite the fact that the topology of FLS is looking simple from the point of view than HFS, the HFS topology provides extra information in their intermediate output variables. Consequently, HFS topology seems to offer a better comprehension and transparent topology in understanding the mapping of input and output process in CPRS.

**Number of rules**

The second criterion of this exploration process is the number of rules. Table 3 presents a summary of the number of rules in FLS and HFS for the CPRS. As can be seen in Table 3, the total number of rules in FLS can be computed using (3) that is 243 rules, based on using three membership functions in each variable. Meanwhile, for the case of HFS, the total number of rules is 36 rules. That is, the summation of rules from each subsystem of HFS, as shown in Table 3.

**Table 3: A summary of the number of rules in FLS and HFS for the CPRS.**

| Number or rules | FLS | HFS |
|---|---|---|
| Subsystem 1 ($FLS_1$) | 243 | 9 |
| Subsystem 2 ($FLS_2$) | - | 9 |
| Subsystem 3 ($FLS_3$) | - | 9 |
| Subsystem 4 ($FLS_4$) | - | 9 |
| Total | 243 | 36 |

From Table 3, we can clearly see that the difference between the number of rules in FLS and HFS is enormous. That is the difference, the number of rules in HFS is approximately about 85% less than FLS. It should be noted that the number of rules can be as an important indicator to see how complex the system it is. For example, as discussed in (Razak et al., 2018), the higher number of rules a system is could indicate the *more complex* the system to design.

**Rules structures**

The third criterion of this exploration is the rules structure. The basic rules structure in FLSs is consist of the if-then rule, which involves two distinct parts: (i) antecedent; and (ii) consequent. The rules are represented in the form:

IF antecedent(s) THEN consequent(s)

Figure 7 shows the rules structure in FLS and HFS for the CPRS. Rules in HFSs are decomposed from FLSs rules into small rules in multiple subsystems, namely $FLS_1$, $FLS_2$, $FLS_3$ and $FLS_4$. By doing this, the rules structure in HFSs is reducing the rule length since it creates a smaller rule for each subsystem. Thus, rules in HFSs are more straightforward than those in FLSs because the number of variables per subsystem is lower (Benítez & Casillas, 2013). Consequently, this may improve the human readability of rule base in HFSs.

| FLS | HFS |
|---|---|
| Rules: | Rules: |
| | **$FLS_1$:** |
| | IF *Q*1 is *Weak* and *Q*2 is *Weak* THEN *Comb_skill 1* is *Weak* |
| | **$FLS_2$:** |
| IF *Q*1 is *Weak* and *Q*2 is *Weak* and *Q*3 is *Weak* and *Q*4 is *Weak* and *Q*5 is *Weak* THEN *Web Programmer* is *NO* | IF *Q*3 is *Weak* and *Q*4 is *Weak* THEN *Comb_skill 2* is *Weak* |
| | **$FLS_3$:** |
| | IF *Comb_skill 1* is *Weak* and *Comb_skill 2* is *Weak* THEN *Comb_skill 3* is *Weak* |
| | **$FLS_4$:** |
| | IF *Comb_skill 3* is *Weak* and *Q*5 is *Weak* THEN *Web Programmer* is *NO* |

**Figure 7. The rules structure in FLS and HFS for the career path recommendation systems (CPRS)**

**Interpretability**

The last criterion of this exploration is interpretability. The criterion is essential, and it is a current debate in the fuzzy community. As discussed earlier, interpretability shows how fuzzy system can be understood by a human being. That is, the fuzzy system is interpretable if people can easily understand it. Consequently, in recent year, the interest of researchers in obtaining more interpretable fuzzy systems has increased.

Table 4: A summary of Interpretability values of FLS and HFS for the CPRS.

| Career path recommendation system (CPRS) | Interpretability index (HFSi) | |
|---|---|---|
| | Fuzzy index | Overall |
| FLS | 0.0642 | **0.0642** |
| HFS: | | |
| • $FLS_1$ | 0.4932 | |
| • $FLS_2$ | 0.4932 | |
| • FLS | 0.4932 | |
| • $FLS_4$ | 0.4932 | |
| | | **0.4932** |

In this section, we used the HFSi index that introduced in (Razak et al., 2017) to measure interpretability of FLS and HFS for the CPRS. Table 4 shows the interpretability measurements of the flat FLS and all subsystems of the HFSs for CPRS using HFSi index. Note that, for this measurement, we used only Fuzzy index joined with the HFSi to measure individual system in HFS.

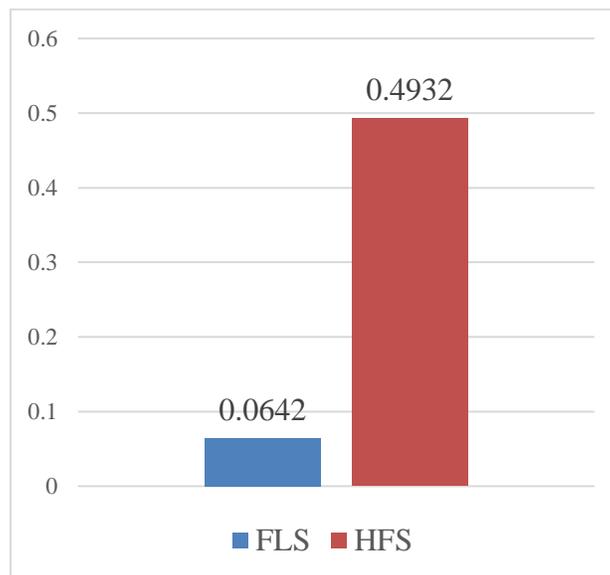

Figure 8. The interpretability values of FLS and HFS for the CPRS

In general, the result of HFSi index values revealed that the HFS is *higher* than FLS for the CPRS, as shown in Figure 8. This could also indicate that the HFS is *more interpretable* than FLS in the context of the CPRS.

## DISCUSSION

This experiment was conducted to explore the use of HFS and compare to flat FLS in the context of career path recommendation system. The experiment was carried based on four key criteria, namely *topology, the number of rules, rules structure* and *interpretability*.

For the first criterion, we have explored and compared the topology of FLS and HFS for the CPRS. From our observation, the result showed the FLS topology is looking simple in showing the mapping of input to the output of the CPRS. It seems possible since FLS topology only consists of one subsystem. Meanwhile, HFS topology consists of several subsystems and located in a different layer of HFS. It was true that from the point of view, the FLS topology looked simple than HFS topology. Despite that, HFS topology provides extra information in their '*intermediate output*', consequently may offer a better comprehensive and transparent topology to be understood by a human being.

For the second criterion, an exploratory study on the number of rules in the FLS and HFS was performed. The outcome showed that the decomposition of FLS to the HFS offers a substantial reduction in the number of rules with about approximately 85%. The finding also indicates that HFS is a practical approach to handle the problem of rule explosion in the flat FLS, especially when dealing with a large number of input variables. Note that in FLS, the number of rules is increased exponentially with the number of input variables (Raju et al., 1991).

In the third criterion, we explored and compared the rules structure in FLS and HFS. From the observation, the outcome showed that by using the same value of '*Weak*' in all input (Q1 - Q5), the output Web programmer produced the same results of '*NO*' in both FLS and HFS for the CPRS. The rules structure in HFS is having a small set of rules in each subsystem, and this may help to improve the readability of a human being. Also, the length of the rules of HFS only consists of the maximum of two antecedents for all rules because of HFS having only two input variables for each subsystem. In contrast, FLS having a maximum of 5 antecedents in their rules, and therefore, it is challenging to be constructed.

For the last criterion, we have investigated and assessed the interpretability of FLS and HFS for the CPRS. That is, the HFSi index was used to measure interpretability of FLS and HFS. The outcome revealed that the HFSi index value produced higher than FLS. This could indicate that the HFS is *more interpretable* than FLS in the context of the CPRS. The outcome seems reasonable, and one may expect that HFS is *more interpretable* than FLS. This is also true since it was confirmed by previous criteria, namely the topology, number of rules and rules structure that showed the HFS has advantages towards interpretability over FLS.

While this exploratory study has shown that HFS is potentially improving the interpretability in FLS, however, it is not cover all aspect of interpretability such as the semantic meaning of fuzzy sets and also intermediate variables. In this context, it is clear that further work is required to establish a comprehensive investigation between FLS and HFS, which covers other interpretability criteria.

## CONCLUSION

In conclusion, we have conducted an initial exploratory study on FLS and HFS for the career path recommendation system. The exploration and comparison were based on four important criteria, namely the topology, number of rules, rules structure and interpretability. Although the explorations are mainly focusing on the four main criteria; however, based on the current evidence, we may conclude that the HFS is promising in improving the interpretability of FLS in the career path recommendation system.

For future work, a further exploratory study on the use of HFS in other recommendation systems will be conducted that incorporates the different criteria of interpretability. Hopefully, the finding obtained will give insight for us to develop the interpretable HFS for the recommendation system.

## ACKNOWLEDGEMENT

We would like to thank the anonymous reviewers for their useful comments and suggestions.